\documentclass[conference]{IEEEtran}
\IEEEoverridecommandlockouts
\usepackage{cite}
\usepackage{amsmath,amssymb,amsfonts}
\usepackage{algorithmic}
\usepackage{graphicx}
\usepackage{textcomp}
\usepackage{xcolor}
\usepackage{booktabs}
\usepackage{subcaption}
\def\BibTeX{{\rm B\kern-.05em{\sc i\kern-.025em b}\kern-.08em
    T\kern-.1667em\lower.7ex\hbox{E}\kern-.125emX}}
\begin{document}

\title{Local Distortion Aware Efficient Transformer Adaptation for Image Quality Assessment}

\author{\IEEEauthorblockN{Kangmin Xu}
\IEEEauthorblockA{\textit{School of Computer Science, WHU, China} \\
xukangmin@whu.edu.cn}
\and
\IEEEauthorblockN{Liang Liao}
\IEEEauthorblockA{\textit{S-lab, NTU, Singapore} \\
liang.liao@ntu.edu.sg}
\and
\IEEEauthorblockN{Jing Xiao}
\IEEEauthorblockA{\textit{School of Computer Science, WHU, China} \\
jing@whu.edu.cn}
\and
\IEEEauthorblockN{Chaofeng Chen}
\IEEEauthorblockA{\textit{S-lab, NTU, Singapore} \\
chaofenghust@gmail.com}
\and
\IEEEauthorblockN{Haoning Wu}
\IEEEauthorblockA{\textit{S-lab, NTU, Singapore} \\
haoning001@e.ntu.edu.sg}
\and
\IEEEauthorblockN{Qiong Yan}
\IEEEauthorblockA{\textit{Sensetime Research and Tetras AI} \\
sophie.yanqiong@gmail.com}
\and
\IEEEauthorblockN{Weisi Lin}
\IEEEauthorblockA{\textit{S-lab, NTU, Singapore} \\
wslin@ntu.edu.sg}
}

\maketitle

\begin{abstract}
Image Quality Assessment (IQA) constitutes a fundamental task within the field of computer vision, yet it remains an unresolved challenge, owing to the intricate distortion conditions, diverse image contents, and limited availability of data. Recently, the community has witnessed the emergence of numerous large-scale pretrained foundation models, which greatly benefit from dramatically increased data and parameter capacities. However, it remains an open problem whether the scaling law in high-level tasks is also applicable to IQA task which is closely related to low-level clues. In this paper, we demonstrate that with proper injection of local distortion features, a larger pretrained and fixed foundation model performs better in IQA tasks. Specifically, for the lack of local distortion structure and inductive bias of vision transformer (ViT), alongside the large-scale pretrained ViT, we use another pretrained convolution neural network (CNN), which is well known for capturing the local structure, to extract multi-scale image features. Further, we propose a local distortion extractor to obtain local distortion features from the pretrained CNN and a local distortion injector to inject the local distortion features into ViT. By only training the extractor and injector, our method can benefit from the rich knowledge in the powerful foundation models and achieve state-of-the-art performance on popular IQA datasets, indicating that IQA is not only a low-level problem but also benefits from stronger high-level features drawn from large-scale pretrained models.
\end{abstract}

\begin{IEEEkeywords}
component, formatting, style, styling, insert
\end{IEEEkeywords}

\section{Introduction}
In the digital era, with millions of images being shared and distributed across various platforms daily, the internet has transformed into a vast repository of visual content. As users exchange and upload images for diverse purposes, spanning from social media interactions to professional applications, ensuring the highest quality and fidelity of these visuals has become highly desirable. Consequently, there has been a substantial increase in the demand for robust image quality assessment (IQA) methods~\cite{yingPatchesPicturesPaQ2PiQ2020, zhangBlindImageQuality2020,sahaReIQAUnsupervisedLearning2023a}. The precise evaluation of image quality holds significant implications, particularly for social media platforms, as it enables them to determine optimal parameter settings for post-upload processing of images, such as resizing, compression, and enhancement, and further ensure a positive user experience, contributing to user satisfaction and engagement.
\begin{figure}[t] 
	\centering
    \includegraphics[width=0.95\linewidth]{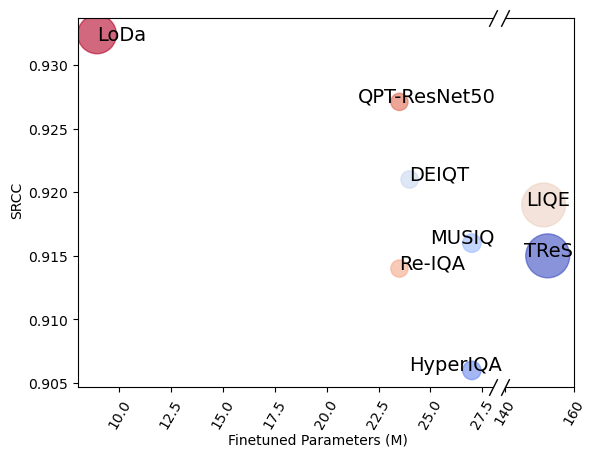}	
    \vspace{-3mm}
	\caption{Comparison among SOTA IQA methods on KonIQ-10k~\cite{hosuKonIQ10kEcologicallyValid2020} dataset, where the size of each spot indicates the model size of the overall network.}
	\label{fig:srcc_fp}
    \vspace{-6mm}
\end{figure}

Amidst the vast volume of data shared on the internet, numerous pretrained large language models~\cite{lewisBARTDenoisingSequencetoSequence2020, brownLanguageModelsAre2020}, vision models~\cite{steinerHowTrainYour2022, ridnikImageNet21KPretrainingMasses2021}, and vision-language models~\cite{radfordLearningTransferableVisual2021, driessPaLMEEmbodiedMultimodal2023} have emerged. Nevertheless, the annotation process for IQA datasets necessitates multiple human annotations for each image, rendering the collection process extremely labor-intensive and financially burdensome. Consequently, the current discipline of IQA suffers from an insufficiency of labeled data, with existing IQA datasets proving inadequate to effectively train large-scale learning models. To address this challenge, a direct approach involves constructing models founded on pretrained Convolutional Neural Networks (CNNs)~\cite{suBlindlyAssessImage2020} or Vision Transformers (ViTs)~\cite{keMUSIQMultiscaleImage2021, qinDataEfficientImageQuality2023a}. Additionally, some studies have proposed the design of IQA-specific pretrained approaches~\cite{sahaReIQAUnsupervisedLearning2023a, zhaoQualityawarePretrainedModels2023a}. Nevertheless, pretraining large models on large datasets requires a considerable investment of time and resources, causing these methodologies to frequently rely on smaller models and datasets, such as ResNet-50~\cite{heDeepResidualLearning2016, zhaoQualityawarePretrainedModels2023a} and ImageNet-1K~\cite{ qinDataEfficientImageQuality2023a, zhaoQualityawarePretrainedModels2023a}.

Vision models have witnessed a progression from EfficientNet based~\cite{phamMetaPseudoLabels2021} architectures (comprising 480M parameters) to Transformer-based counterparts~\cite{yuCoCaContrastiveCaptioners2022} with an incredible parameter count of 2,100M, and more recently, they have risen to unprecedented scales, encompassing 22B parameters~\cite{dehghaniScalingVisionTransformers2023} and 562B~\cite{driessPaLMEEmbodiedMultimodal2023}, with expectations for further growth. Given the magnitude of such large models, traditional pretraining and full-finetuning approaches prove exceptionally challenging, as they necessitate the entire process of each model for every specific task. In light of this, and drawing inspiration from efficient model adaptation techniques in natural language processing (NLP)~\cite{zhouComprehensiveSurveyPretrained2023}, a variety of visual tuning methods~\cite{jiaVisualPromptTuning2022a, zhangNeuralPromptSearch2022} have emerged, enabling the adaptation of pretrained vision or visual-language models to downstream tasks. This practice is different from the operating procedure of transfer learning that either fully fine-tunes the whole model or just fine-tunes the task head~\cite{zhuangComprehensiveSurveyTransfer2021}. As such, whether or not IQA models can leverage shared parameter weights (typically interpreted as the knowledge of pre-trained models) from large-scale pretrained models to improve performance remains of the greatest significance and interest.

In this work, we make the first attempt to efficiently adapt large-scale pretrained models to IQA tasks, namely LOcal Distortion Aware efficient transformer adaptation (LoDa). The majority of large-scale pretrained models~\cite{ridnikImageNet21KPretrainingMasses2021,  driessPaLMEEmbodiedMultimodal2023} are grounded in the Transformer architecture~\cite{vaswaniAttentionAllYou2017}, which is powerful for modeling non-local dependencies~\cite{golestanehNoReferenceImageQuality2022, qinDataEfficientImageQuality2023a}, but it is weak for local structure and inductive bias~\cite{yuanTokenstoTokenViTTraining2021}. However, IQA is highly reliant on both local and non-local features~\cite{golestanehNoReferenceImageQuality2022, qinDataEfficientImageQuality2023a}. In addition, as the human visual system captures an image in a multi-scale fashion, previous works~\cite{keMUSIQMultiscaleImage2021, golestanehNoReferenceImageQuality2022} have also shown the benefit of using multi-scale features extracted from CNNs feature maps at different depths for IQA. With the obtained insights, we propose to inject multi-scale features extracted by CNNs into ViT, thereby enriching its representation with local distortion features and inductive bias. 

Specifically, we feed input images into both a pretrained CNN and a large-scale pretrained ViT, yielding a set of multi-scale features. Then we employ convolution and average pooling processes to collect multi-scale distortion features while discarding redundant data from the multi-scale features. However, the process of infusing these multi-scale features into ViT is not straightforward. Indeed, although we can manipulate and reshape the multi-scale features to mirror the shape of ViT tokens and simply merge them, it is crucial to acknowledge that an image token within ViT corresponds to a \(16 \times 16\) patch extracted from the original image, which might not align with the scale of the multi-scale features. To this end, we introduce the cross-attention mechanism, allowing us to query features resembling the image token of ViT from the multi-scale features. These queried features are subsequently fused with the image tokens, ensuring a seamless and meaningful integration of the distortion-related data.

Furthermore, considering the substantial channel dimension of the large-scale pretrained vision transformer (768 for ViT-B), it is imperative to address potential issues stemming from employing this dimension directly in the context of cross-attention. It could lead to an overwhelming number of parameters and computational overhead, which is inconsistent with the principles of efficient model adaptation. Taking inspiration from the concept of adapters in the field of NLP~\cite{houlsbyParameterEfficientTransferLearning2019}, we propose to down-project ViT tokens and multi-scale distortion features to a smaller dimension, which serves to mitigate parameter increase and computational demands. 
In general, the contributions of this paper can be summarized in three-folds: 
\begin{itemize}
    \item We make the first attempt to efficiently adapt large-scale pretrained models to IQA tasks. We leverage the knowledge of large-scale pretrained models to develop an IQA model that only introduces small trainable parameters to alleviate the scarcity of training data. 
    \item We embed supplementary multi-scale features obtained from pretrained CNNs into large-scale pretrained ViTs. With proper local distortion injection, a larger pretrained backbone could show better IQA performance.
    \item Extensive experiments on seven IQA benchmarks show that our method significantly outperforms other counterparts with much less trainable parameters, indicating the effectiveness and generalization ability of our methods.
\end{itemize}

\section{Related Work}
\subsection{Deep Learning Based Image Quality Assessment}

 \begin{figure*}[] 
	\centering
	\includegraphics[width=0.98\textwidth]{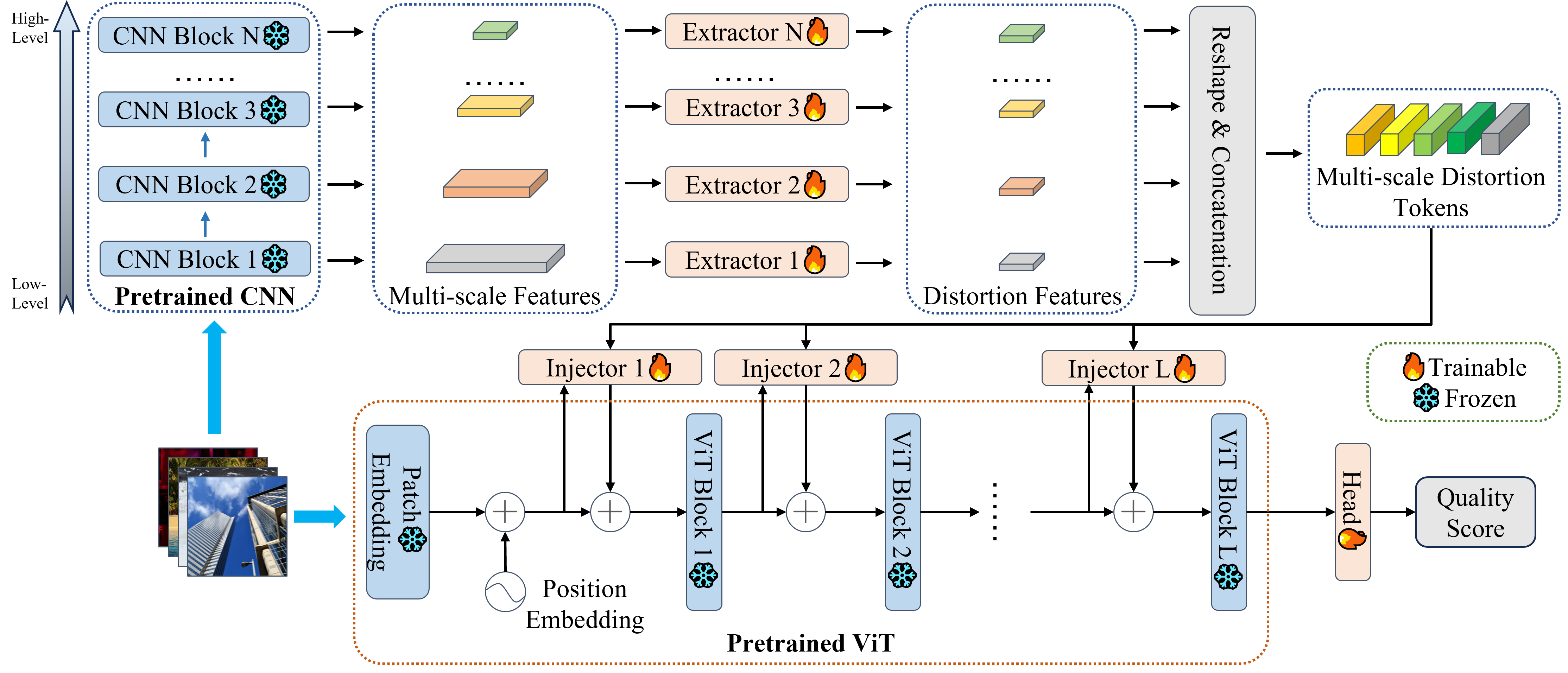}
 \vspace{-3mm}
	\caption{Framework overview of the proposed LoDa}
	\label{fig:arch}
 \vspace{-3mm}
\end{figure*}

By the success of deep learning in many computer vision tasks, different approaches utilize deep learning for IQA: early CNN-based~\cite{suBlindlyAssessImage2020, yingPatchesPicturesPaQ2PiQ2020, zhangBlindImageQuality2020} and recently transformer-based methods~\cite{ golestanehNoReferenceImageQuality2022, qinDataEfficientImageQuality2023a}. Modern CNN-based models commonly posit that initial stages within the network encapsulate low-level spatial characteristics, whereas subsequent stages are indicative of higher-level semantic features~\cite{ huNoReferenceQualityMetric2023}. Based on this, Su \textit{et al.}~\cite{suBlindlyAssessImage2020} put forth a method wherein multi-scale features and semantic features are extracted from images using the ResNet architecture~\cite{heDeepResidualLearning2016}. Subsequently, they endeavor to capture local distortion information from the multi-scale features and generate weights utilizing semantic features to serve as a quality prediction target network. Lastly, the target network assimilates the aggregated local distortion features as input to predict image quality.

Although CNNs capture the local structure of the image, they are well known for missing to capture non-local information and having strong locality bias. On the contrary, Vision Transformer (ViT)~\cite{dosovitskiyImageWorth16x162021} has strong capability in modeling the non-local dependencies among features of the image, thus transformer-based methods demonstrate great potential in dealing with the image quality assessment. Golestaneh, Dadsetan, and Kitani~\cite{golestanehNoReferenceImageQuality2022} proposed a method that utilizes CNNs to extract the perceptual features as inputs to the Transformer encoder. Ke \textit{et al.}~\cite{keMUSIQMultiscaleImage2021} and Qin \textit{et al.}~\cite{qinDataEfficientImageQuality2023a} directly send image patches as inputs to the Transformer encoder.

\subsection{Large-scale Pretrained Models}
Recently, the parameter capacities of vision models are undergoing a rapid expansion, scaling from 480M parameters of EfficientNet-based~\cite{phamMetaPseudoLabels2021} to 22B parameters of Transformer-based counterparts~\cite{dehghaniScalingVisionTransformers2023}. As a consequence, their requisition for training data and training techniques is similarly expanding. In terms of this, these models are commonly trained on large-scale labeled datasests~\cite{ridnikImageNet21KPretrainingMasses2021, zhaiScalingVisionTransformers2022} in a supervised or self-supervised manner. Moreover, some works~\cite{driessPaLMEEmbodiedMultimodal2023} adopt large-scale multi-modal data (\textit{e.g.}, image-text pairs) for training, which leads to even more powerful visual representations. In this work, we could take advantage of these well pretrained image models and adapt them efficiently to solve IQA tasks.

\subsection{Efficient Model Adaptation}
In the field of NLP, efficient model adaptation techniques involve adding or modifying a limited number of parameters of the model, as limiting the dimension of the optimization problem can prevent catastrophic forgetting. Conventional arts~\cite{suBlindlyAssessImage2020} typically adopt full-tuning in the downstream tasks. Rare attention has been drawn to the field of efficient adaptation, especially in the field of vision Transformers. However, with the surge of large-scale pretrained models, the conventional paradigm is inevitably limited by the huge computational burden, thus some works~\cite{jiaVisualPromptTuning2022a, zhangNeuralPromptSearch2022} migrate the efficient model adaptation techniques that appeared in NLP to CV. 

Due to the paucity of labeled data for training, IQA methods are unable to realize their full potential. Previous works~\cite{suBlindlyAssessImage2020, golestanehNoReferenceImageQuality2022} commonly full-finetune the whole network trained on ImageNet-1K, but the model and data are insufficiently large. In this work, we propose employing efficient model adaptation techniques to adapt large-scale pretrained models to IQA tasks. 

\section{The Proposed Method}
\subsection{Overall Architecture}
In an effort to further improve the efficiency of pretrained model adaptation and customize it for IQA tasks, we devise a transformer-based adaptation efficient framework, namely LOcal Distortion Aware efficient transformer adaptation (LoDa). As depicted in Figure~\ref{fig:arch}, the framework of LoDa is composed of two components. The initial component incorporates a large-scale pretrained Vision Transformer (ViT)~\cite{dosovitskiyImageWorth16x162021}. The second component comprises a pretrained CNN tasked with extracting multi-scale features from the input image. Moreover, it integrates a local distortion extractor responsible for capturing local distortion features from the extracted multi-scale features. Subsequently, a local distortion-aware injector is employed to procure corresponding local distortion tokens, which are similar to tokens of the ViT model, and then infuse them for the later stage.

Specifically, upon receiving an input image, our process initiates by directing it to a pretrained CNN to extract multi-scale features. Subsequently, these mult-scale feature maps are individually routed into separate local distortion extractors, generating distinct local distortion features. These local distortion features are then reshaped and concatenated to create multi-scale distortion tokens for later interaction. Simultaneously, the input image is further input into the patch embedding layer of the pretrained ViT. Here, the image undergoes division into non-overlapping \(16 \times 16\) patches, which are then flattened and transformed into $D$-dimensional tokens by the patch embedding layer. Subsequently, these tokens are added with position embeddings. After this process, the tokens, acting as queries, are coupled with the multi-scale distortion tokens and are subjected to cross-attention. This results in the extraction of similar local distortion features from the multi-scale distortion tokens, which are subsequently injected into the tokens of the ViT, thereby enhancing the distortion-related information encompassed by these tokens. Following this, the tokens, along with the augmented distortion features, traverse through $L$ transformer encoder layers and cross-attention blocks. Finally, the CLS token acquired from the ViT serves as the input to the quality regressor, enabling the derivation of the final quality score.

It is noteworthy that during adaptation, only the local distortion extractor modules, local distortion aware injectors and the head are trainable, but the weights of the pretrained ViT encoder and pretrained CNN are frozen.

\begin{figure}[t]
	\centering
        \begin{subfigure}{\linewidth}
            \centering
		    \includegraphics[width=\linewidth]{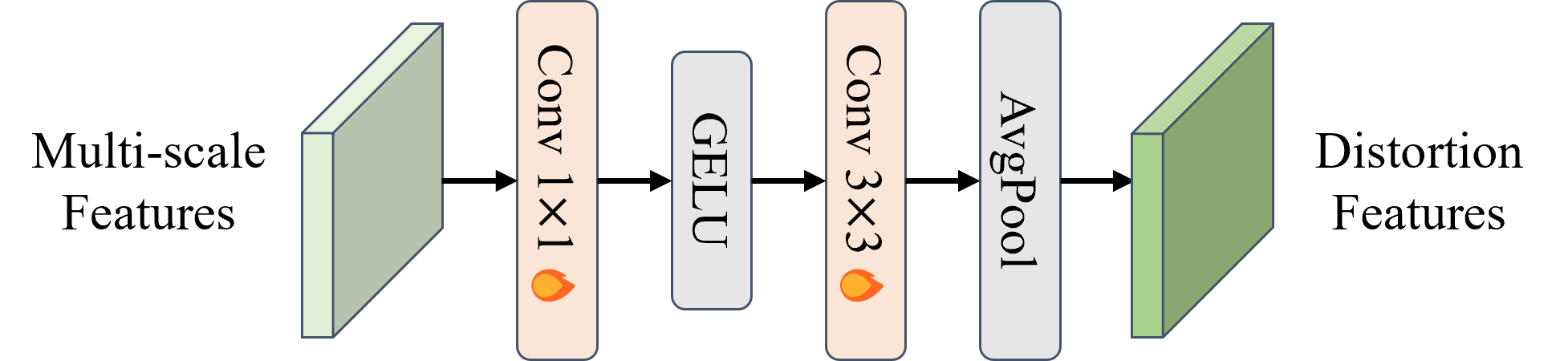}
            \caption{Local Distortion Extractor}
            \label{fig:lde}
        \end{subfigure}
        \vfill
        \vspace{1mm}
        \begin{subfigure}{\linewidth}
            \centering
		    \includegraphics[width=\linewidth]{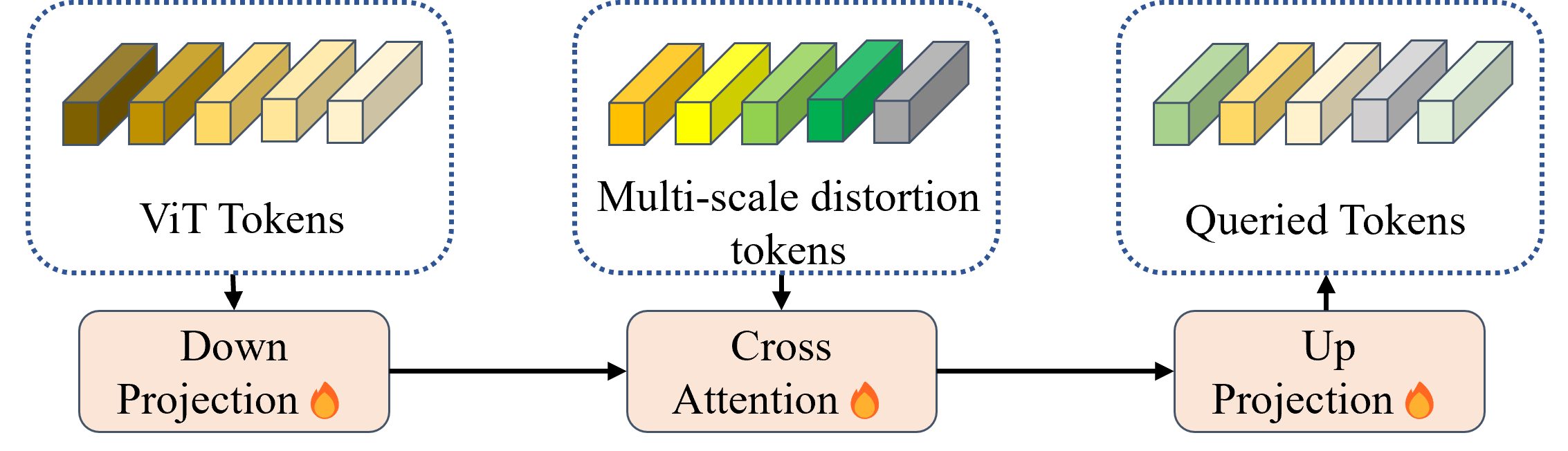}
            \caption{Local Distortion Injector}
            \label{fig:ldi}
        \end{subfigure}
    \vspace{-2mm}
    \caption{The architecture of the proposed local distortion extractor and injector.}
    \label{fig:lde-ldi}
    \vspace{-3mm}
\end{figure}

\subsection{Local Distortion Extractor}
The majority of large-scale pretrained models~\cite{ridnikImageNet21KPretrainingMasses2021, steinerHowTrainYour2022, yuCoCaContrastiveCaptioners2022, driessPaLMEEmbodiedMultimodal2023} are grounded in the Transformer architecture~\cite{vaswaniAttentionAllYou2017}, renowned for its robust capacity to model non-local dependencies among perceptual features within an image. However, these models exhibit a weak inductive bias. Conversely, CNNs excel at capturing the local structure of an image, exhibiting a strong locality bias, but they falter in capturing non-local information~\cite{golestanehNoReferenceImageQuality2022, qinDataEfficientImageQuality2023a}. Nonetheless, IQA is highly reliant on both local and non-local features~\cite{golestanehNoReferenceImageQuality2022, qinDataEfficientImageQuality2023a}. In the absence of abundant labeled data~\cite{sahaReIQAUnsupervisedLearning2023a, zhaoQualityawarePretrainedModels2023a}, the adaptation of large-scale pretrained models to IQA may suffer from a deficiency in local structure and inductive bias. This deficiency can, however, be mitigated by leveraging the capabilities of CNNs~\cite{siInceptionTransformer2022}. In light of these considerations, we propose the exploitation of the local structure and inductive bias derived from pretrained CNNs to strengthen the adaptation of large-scale pretrained models for IQA without altering their original architecture.

As shown in Figure~\ref{fig:arch}, with the given input image \(I \in \mathbb{R}^{H \times W \times C}\), the pretrained CNN will output a set of multi-scale features \(F^j \in \mathbb{R}^{b \times c_j \times m_j \times n_j}\), where \(j\) denotes the $j$-th block of CNN, \(b\) denotes the batch size and \(c_j\), \(m_j\) and \(n_j\) denote the channel size, width, and height of the $j$-th features, respectively. The reason why we extract multi-scale features is that semantic features extracted from the last layer merely represent holistic image content~\cite{suBlindlyAssessImage2020}. In order to capture local distortions in the real world, we propose to extract multi-scale features \(F^j\) through a local distortion extractor. As illustrated in Figure~\ref{fig:lde}, we use sequential trainable \(1 \times 1\) and \(3 \times 3\) convolutional layers to project them into equal dimensions and further extract image quality-related local distortion features and inductive bias from the multi-scale features. Since the initial features of pretrained CNNs have relatively large dimensions, to keep adaptation efficiency, we use average pooling to pool the extracted features into a smaller size. Let \(\bar{F}^j \in \mathbb{R}^{b, c, m, n}\) denote the output feature after sending \(F^j\) to the convolutions and pooling. Next, we flatten and concatenate \(\bar{F}^j\) and obtain the multi-scale distortion tokens \(F_{msd} \in \mathbb{R}^{b, \sum_j (m \times n), c}\), as the input for the local distortion injector. 

\subsection{Local Distortion Injector}
A direct approach to infusing multi-scale distortion tokens into tokens of large-scale pretrained ViT models involves a simple addition of the features with the tokens. Nevertheless, it should be noted that an image token in ViT corresponds to a \(16 \times 16\) patch of the original image, which might not align with the scale of the multi-scale distortion features. To address this misalignment, we introduce to use cross-attention mechanism, which enables to query features akin to the image token of ViT from the multi-scale distortion features. Subsequently, the queried features are adeptly combined with the image tokens, ensuring a coherent and effective integration of the distortion information.

As illustrated in Figure~\ref{fig:ldi}, after passing the input image \(I\) to large-scale pretrained ViT, assuming that \(F_{vit}^i \in \mathbb{R}^{b, l, d}\) denote the token of \(i^{th}\) block of the ViT (including CLS token and image token). We take \(F_{vit}^i\) as query \(Q_i\) and multi-scale distortion tokens \(F_{msd}\) as key \(K_i\) and value \(V_i\) of multi-head cross-attention (MHCA) to obtain multi-scale distortion tokens that are similar to \(F_{vit}^i\) from \(F_{msd}\): 
\begin{equation}
    \bar{F}_{msd}^i = MHCA(Q_i, K_i, V_i) + Q_i.
    \label{eq:1}
\end{equation}

Then, the queried multi-scale distortion tokens are added with ViT tokens \(F_{vit}^i\), which can be written as Eqn.~\ref{eq:2}: 
\begin{equation}
    \bar{F}_{vit}^i = F_{vit}^i + s^i \times \bar{F}_{msd}^i,
    \label{eq:2}
\end{equation}
where \(s^i\) represents a trainable vector designed to strike a balance between the output of the attention layer and the input feature \(F_{vit}^i\). To facilitate this balance, \(s^i\) is initialized to a value close to 0. This specific initialization strategy ensures that the feature distribution of \(F_{vit}^i\) remains unchanged despite the injection of queried multi-scale distortion features, thereby allowing for more effective utilization of the pretrained weights of ViT in the adaptation process.

Due to the channel dimension of the large-scale pretrained vision transformer being relatively large (768 for ViT-B), directly using this for extra MHCA will bring a tremendous amount of parameters and computational overhead, which is not consistent with efficient model adaptation. Inspired by adapter~\cite{houlsbyParameterEfficientTransferLearning2019} in NLP, we propose to down project ViT token \(F_{vit}^i\) and multi-scale distortion features \(F_{msd}\) to a smaller dimension \(r\), 
\begin{equation}
    \tilde{F}_{vit}^i = f(F_{vit}^i),
    \label{eq:3}
\end{equation}
\begin{equation}
    \tilde{F}_{msd} = f(F_{msd}),
    \label{eq:4}
\end{equation}
where \(f\) is a trainable MLP layer, performs the projection of ViT token \(F_{vit}^i\) and multi-scale distortion features \(F_{msd}\) into \(\tilde{F}_{vit}^i \in \mathbb{R}^{b, l, r}\) and \(\tilde{F}_{msd} \in \mathbb{R}^{b, \sum_j (m \times n), r}\), respectively. Notably, it is \(\tilde{F}_{vit}^i\) and \(\tilde{F}_{msd}\) that take on the roles of query \(Q_i\), key \(K_i\) and value \(V_i\) within MHCA, instead of \(F_{vit}^i\) and \(F_{msd}\). Lastly, we up-project the result from cross-attention by a trainable MLP layer into the dimension of ViT tokens. 

\subsection{IQA Regression}
With the output CLS token of ViT, we feed it into a single-layer regressor head to obtain the quality score. A PLCC-induced loss is employed for training. Assuming there are \(m\) images on the training batch and the predicted quality scores \(\tilde{y} = \{\tilde{y}_1, \tilde{y}_2, \dots, \tilde{y}_m\}\) and corresponding label \(y = \{y_1, y_2, \dots, y_m\}\), the PLCC-induced loss is defined as: 
\begin{equation}
    \mathcal{L}_{p l c c}=\left(1-\frac{\sum_{i=1}^{m}\left(\tilde{y}_i-\tilde{a}\right)\left(y_{i}-a\right)}{\sqrt{\sum_{i=1}^{m}\left(\tilde{y}_i-\tilde{a}\right)^{2} \sum_{i=1}^{m}\left(y_{i}-a\right)^{2}}}\right) / 2
    \label{eq:plcc_loss}
\end{equation}
where \(\tilde{a}\) and \(a\) are the mean values of \(\tilde{y}\) and \(y\), respectively.

\begin{table*}[]
	\centering
 \renewcommand\tabcolsep{4pt}
	\resizebox{\textwidth}{!}{
		\begin{tabular}{lcccccc||cccccccc}
			\toprule[1.5pt] & \multicolumn{2}{c}{ LIVE } & \multicolumn{2}{c}{ TID2013 } & \multicolumn{2}{c||}{ KADID-10k } & \multicolumn{2}{c}{ LIVEC } & \multicolumn{2}{c}{ KonIQ-10k } & \multicolumn{2}{c}{ SPAQ } & \multicolumn{2}{c}{ FLIVE } \\
			\cmidrule{2-15}
			\Large{Method} & SRCC & PLCC & SRCC & PLCC & SRCC & PLCC & SRCC & PLCC & SRCC & PLCC & SRCC & PLCC & SRCC & PLCC \\
			\midrule[1pt] 
			ILNIQE & 0.902 & 0.906 & 0.521 & 0.648 & 0.534 & 0.558 & 0.508 & 0.508 & 0.523 & 0.537 & 0.713 & 0.712 & 0.294 & 0.332 \\
			BRISQUE & 0.929 & 0.944 & 0.626 & 0.571 & 0.528 & 0.567 & 0.629 & 0.629 & 0.681 & 0.685 & 0.809 & 0.817 & 0.303 & 0.341 \\
			WaDIQaM-NR & 0.960 & 0.955 & 0.835 & 0.855 & 0.739 & 0.752 & 0.682 & 0.671 & 0.804 & 0.807 & - & - & 0.455 & 0.467 \\
			DB-CNN & 0.968 & 0.971 & 0.816 & 0.865 & 0.851 & 0.856 & 0.851 & 0.869 & 0.875 & 0.884 & 0.911 & 0.915 & 0.545 & 0.551 \\
			TIQA & 0.949 & 0.965 & 0.846 & 0.858 & 0.850 & 0.855 & 0.845 & 0.861 & 0.892 & 0.903 & - & - & 0.541 & 0.581 \\
			MetaIQA & 0.960 & 0.959 & 0.856 & 0.868 & 0.762 & 0.775 & 0.835 & 0.802 & 0.887 & 0.856 & - & - & 0.540 & 0.507 \\
			P2P-BM & 0.959 & 0.958 & 0.862 & 0.856 & 0.840 & 0.849 & 0.844 & 0.842 & 0.872 & 0.885 & - & - & 0.526 & 0.598 \\
			HyperIQA ({\it 27M}) & 0.962 & 0.966 & 0.840 & 0.858 & 0.852 & 0.845 & 0.859 & 0.882 & 0.906 & 0.917 & 0.911 & 0.915 & 0.544 & 0.602 \\
			MUSIQ ({\it 27M}) & 0.940 & 0.911 & 0.773 & 0.815 & 0.875 & 0.872 & 0.702 & 0.746 & 0.916 & 0.928 & 0.918 & 0.921 & 0.566 & 0.661 \\
			TReS ({\it 152M}) & 0.969 & 0.968 & 0.863 & 0.883 & 0.859 & 0.858 & 0.846 & 0.877 & 0.915 & 0.928 & - & - & 0.544 & 0.625 \\
			DEIQT (24M) & \textbf{0.980} & \textbf{0.982} & \textbf{0.892} & \textbf{0.908} & 0.889 & 0.887 & 0.875 & 0.894 & 0.921 & 0.934 & 0.919 & 0.923 & 0.571 & 0.663 \\
			LIQE ({\it 151M}) & 0.970 & 0.951 & - & - & \textbf{0.930} & \textbf{0.931} & \textbf{0.904} & \textbf{0.910} & 0.919 & 0.908 & - & - & - & - \\
			Re-IQA ({\it 24M}) & 0.970 & 0.971 & 0.804 & 0.861 & 0.872 & 0.885 & 0.840 & 0.854 & 0.914 & 0.923 & 0.918 & 0.925 & - & - \\
            QPT-ResNet50 ({\it 24M}) & - & - & - & - & - & - & \textbf{0.895} & \textbf{0.914} & \textbf{0.927} & \textbf{0.941} & \textbf{0.925} & \textbf{0.928} & \textbf{0.575} & \textbf{0.675} \\
			\midrule[1pt]
			LoDa ({\it 9M}) & \textbf{0.975} & \textbf{0.979} & \textbf{0.869} & \textbf{0.901} & \textbf{0.931} & \textbf{0.936} & 0.876 & 0.899 & \textbf{0.932} & \textbf{0.944} & \textbf{0.925} & \textbf{0.928} & \textbf{0.578} & \textbf{0.679} \\
			\bottomrule[1.5pt]
	\end{tabular}}
 \vspace{-2mm}
	\caption{Performance comparison measured by medians of SRCC and PLCC, where bold entries indicate the top two results.}
\label{tab:sota}
 \vspace{-4mm}
\end{table*}

\section{Experiments}
\subsection{Experimental Setting}
\subsubsection{Datasets}
Our method is evaluated on seven classical IQA datasets, including three synthetic datasets of LIVE~\cite{sheikhStatisticalEvaluationRecent2006}, TID2013~\cite{ponomarenkoImageDatabaseTID20132015}, KADID-10k~\cite{linKADID10kLargescaleArtificially2019} and four authentic datasets of LIVEC~\cite{ghadiyaramMassiveOnlineCrowdsourced2015}, KonIQ-10k~\cite{hosuKonIQ10kEcologicallyValid2020}, SPAQ~\cite{fangPerceptualQualityAssessment2020}, FLIVE~\cite{yingPatchesPicturesPaQ2PiQ2020}. For the synthetic datasets, they contain a few pristine images which are synthetically distorted by various distortion types, such as JPEG compression and Gaussian blurring. LIVE contains 799 synthetically distorted images with 5 distortion types. TID2013 and KADID-10k consist of 3000 and 10125 synthetically distorted images involving 24 and 25 distortion types, respectively. For the authentic datasets, LIVEC consists of 1,162 images with diverse authentic distortions captured by mobile devices. KonIQ-10k contains 10,073 images which are selected from YFCC-100M and the selected images cover a wide and uniform range of distortions such as brightness colorfulness, contrast, noise, sharpness, etc. SPAQ consists of 11,125 images captured by different mobile devices, covering a large variety of scene categories. FLIVE is the largest in-the-wild IQA dataset by far, which contains 39,810 real-world images with diverse contents, sizes, and aspect ratios. 

\subsubsection{Evaluation Criteria}
Spearman’s rank order correlation coefficient (SRCC) and Pearson’s linear correlation coefficient (PLCC) are employed to measure prediction monotonicity and prediction accuracy. The higher value indicates better performance. For PLCC, a logistic regression correction is also applied according to VQEG~\cite{antkowiakFinalReportVideo2000}. 

\begin{table}[t]
	\centering
 \renewcommand\tabcolsep{6.5pt}
	\resizebox{\columnwidth}{!}{ 
		\begin{tabular}{lcccccc}
			\toprule Training & \multicolumn{2}{c}{ FLIVE } & LIVEC & KonIQ  \\
			\midrule[0.25pt] Testing & KonIQ & LIVEC & KonIQ & LIVEC  \\
			\midrule[1pt]
			DBCNN  & 0.716 & 0.724 & 0.754 & 0.755  \\
			P2P-BM  & 0.755 & 0.738 & 0.740 & 0.770 \\
			HyperIQA  & 0.758 & 0.735 & 0.772 & 0.785 \\
			TReS  & 0.713 & 0.740 & 0.733 & 0.786 & \\
			DEIQT & 0.733 & 0.781 & 0.744 & 0.794 \\
			\midrule[1pt]
			LoDa & \textbf{0.763} & \textbf{0.805} & \textbf{0.745} & \textbf{0.811} \\
			\bottomrule
		\end{tabular}
	}
 \vspace{-2mm}
	\caption{SRCC on the cross datasets validation. The best performances are highlighted with \textbf{boldface}.}
	\label{tab:cross}
 \vspace{-3mm}
\end{table}

\subsubsection{Implementation Details}
We implement the model by PyTorch and conduct training and testing on an NVIDIA RTX 4090 GPU. We resize the smaller edge of images to 384, randomly crop an input image into multiple image patches with a resolution of \(224 \times 224\), and horizontally and vertically augment them randomly to increase the number of data for training~\cite{zhangBlindImageQuality2020}. Particularly, the number of patches for training is determined depending on the size of the dataset, \textit{i.e.}, 1 for FLIVE, 3 for KonIQ-10k, and 5 for LIVEC, the number of patches for testing is 15 for all datasets, and patches inherit quality scores from the source image. We create our model based on the ViT-B pretrained on ImageNet-21k with an image size of \(224 \times 224\) and patch size of \(16 \times 16\). We use ResNet50~\cite{heDeepResidualLearning2016} pretrained on ImageNet-1k for the CNN backbone and extract feature maps of the last four blocks as multi-scale features. we use average pooling to pool multi-scale features into a spatial size of \(7 \times 7\). The dimension after the down projection is 64 and the number of heads used for cross-attention is 4. 

We use AdamW optimizer with a weight decay of 0.01 and mini-batch size of 128. The learning rate was initialized with 0.0003 and decayed by the cosine annealing strategy. All experiments are trained for 10 epochs. By default, we select the evaluation of the last epoch. For each dataset, 80\% images were used for training and the remaining 20\% images were utilized for testing. We repeated this process 10 times for all experiments to mitigate the performance bias and the medians of SRCC and PLCC were reported.

\subsection{Comparisons with the State-of-the-art Methods}
The performance comparison over the State-of-the-art (SOTA) BIQA methods is shown in Table~\ref{tab:sota}. Our model outperforms the existing SOTA methods~\cite{zhangFeatureEnrichedCompletelyBlind2015, mittalNoReferenceImageQuality2012, bosseDeepNeuralNetworks2018, zhangBlindImageQuality2020, youTransformerImageQuality2021, zhuMetaIQADeepMetaLearning2020, yingPatchesPicturesPaQ2PiQ2020, suBlindlyAssessImage2020, keMUSIQMultiscaleImage2021, golestanehNoReferenceImageQuality2022, qinDataEfficientImageQuality2023a, zhangBlindImageQuality2023, sahaReIQAUnsupervisedLearning2023a, zhaoQualityawarePretrainedModels2023a} by a significant margin on these datasets of both synthetically and authentically distorted images. Since images on various datasets span a wide variety of image contents and distortion types, it is still challenging to consistently achieve the leading performance on all of them. 

\begin{table}[t]
    \centering
    \footnotesize
 \renewcommand\tabcolsep{8pt}
	\resizebox{\columnwidth}{!}{ 
    \begin{tabular}{c|cc|cc}
    \toprule
     & \multicolumn{2}{c|}{KADID-10k} & \multicolumn{2}{c}{KonIQ-10k}  \\
    Pre-train & SRCC & PLCC & SRCC & PLCC \\
    \midrule
    MAE & 0.917 & 0.924 & 0.927 & 0.938 \\
    Multi-Modal & 0.897 & 0.902 & 0.909 & 0.923 \\
    ImageNet-1K & 0.912 & 0.920 & 0.920 & 0.933 \\
    ImageNet-21K & \textbf{0.931} & \textbf{0.936} & \textbf{0.932} & \textbf{0.944} \\
    \bottomrule
    \end{tabular}}
 \vspace{-2mm}
    \caption{Impact of large-scale pretrained models, using different methods and datasets pretrained models. }
    \label{tab:pre-train}
    \vspace{-0.4cm}
\end{table}

Specifically, ours surpass traditional methods (\textit{e.g.}, ILNIQE~\cite{zhangFeatureEnrichedCompletelyBlind2015} and BRISQUE~\cite{mittalNoReferenceImageQuality2012}) and earlier learning-based methods (\textit{e.g.}, TIQA~\cite{youTransformerImageQuality2021} and HyperIQA~\cite{suBlindlyAssessImage2020} by a large margin. For LIQE~\cite{zhangBlindImageQuality2023} that utilized a large-scale pretrained vision-language model, multi-task labels, and full fine-tuning on multiple datasets simultaneously, LoDa still outperforms on both large synthetical and authentical datasets, \textit{i.e.}, KADID10k and KonIQ10k. Compared with current SOTA methods that required extra pertaining (\textit{e.g.}, DEIQT~\cite{qinDataEfficientImageQuality2023a}, Re-IQA~\cite{sahaReIQAUnsupervisedLearning2023a} and QPT-ResNet50~\cite{zhaoQualityawarePretrainedModels2023a}), LoDa obtains competitive or higher results, showing the powerful effectiveness of adaptation of large-scale pretrained models. Correspondingly, the top performance on the largest synthetical datasets KADID-10k confirms the superiority of fusing the multi-scale distortion features from CNN into ViT model. 

\subsection{Cross-Dataset Evaluation}
We further compare the generalizability of LoDa against competitive BIQA models in a cross-dataset setting following~\cite{qinDataEfficientImageQuality2023a}. Training is performed on one specific dataset, and testing is performed on a different dataset without any finetuning or parameter adaptation. The experimental results in terms of the medians of SRCC on four datasets are reported in Table~\ref{tab:cross}. As observed, LoDa achieves the best performance on all datasets. These results manifest the strong generalization capability of LoDa. 

\begin{table}[t]
    \centering
    \footnotesize
 \renewcommand\tabcolsep{9pt}
	\resizebox{\columnwidth}{!}{ 
    \begin{tabular}{c|cc|cc}
    \toprule
     & \multicolumn{2}{c|}{KADID-10k} & \multicolumn{2}{c}{KonIQ-10k}  \\
    Backbone & SRCC & PLCC & SRCC & PLCC \\
    \midrule
    ViT-T & 0.892 & 0.900 & 0.914 & 0.926 \\
    ViT-S & 0.915 & 0.922 & 0.928 & 0.939 \\
    ViT-B & \textbf{0.931} & \textbf{0.936} & \textbf{0.932} & \textbf{0.944} \\
    \bottomrule
    \end{tabular}}
 \vspace{-2mm}
    \caption{Impact of large-scale pretrained model sizes. }
    \label{tab:model-size}
    \vspace{-0.3cm}
\end{table}

\begin{table}[t]
    \centering
    \footnotesize
 \renewcommand\tabcolsep{5pt}
	\resizebox{\columnwidth}{!}{ 
    \begin{tabular}{c|cc|cc}
    \toprule
     & \multicolumn{2}{c|}{KADID-10K} & \multicolumn{2}{c}{KonIQ-10K}  \\
    Fine-tuning Methods & SRCC & PLCC & SRCC & PLCC \\
    \midrule
    ViT (Linear Probe) & 0.676 & 0.701 & 0.796 & 0.833 \\
    ViT (Full fine-tune) & 0.889 & 0.899 & 0.874 & 0.891 \\
    Adapter-ViT & 0.914 & 0.920 & 0.926 & 0.939 \\
    LoRA-ViT & 0.913 & 0.921 & 0.921 & 0.934 \\
    VPT-ViT & 0.889 & 0.900 & 0.919 & 0.932 \\
    LoDa & \textbf{0.931} & \textbf{0.936} & \textbf{0.932} & \textbf{0.944} \\
    \bottomrule
    \end{tabular}}
 \vspace{-2mm}
    \caption{Comparisons with different fine-tuning methods. }
    \label{tab:ema}
    \vspace{-0.4cm}
\end{table}

\begin{figure}[t]
\centering
    \begin{subfigure}{0.29\linewidth}
        \centering
        \includegraphics[width=\linewidth]{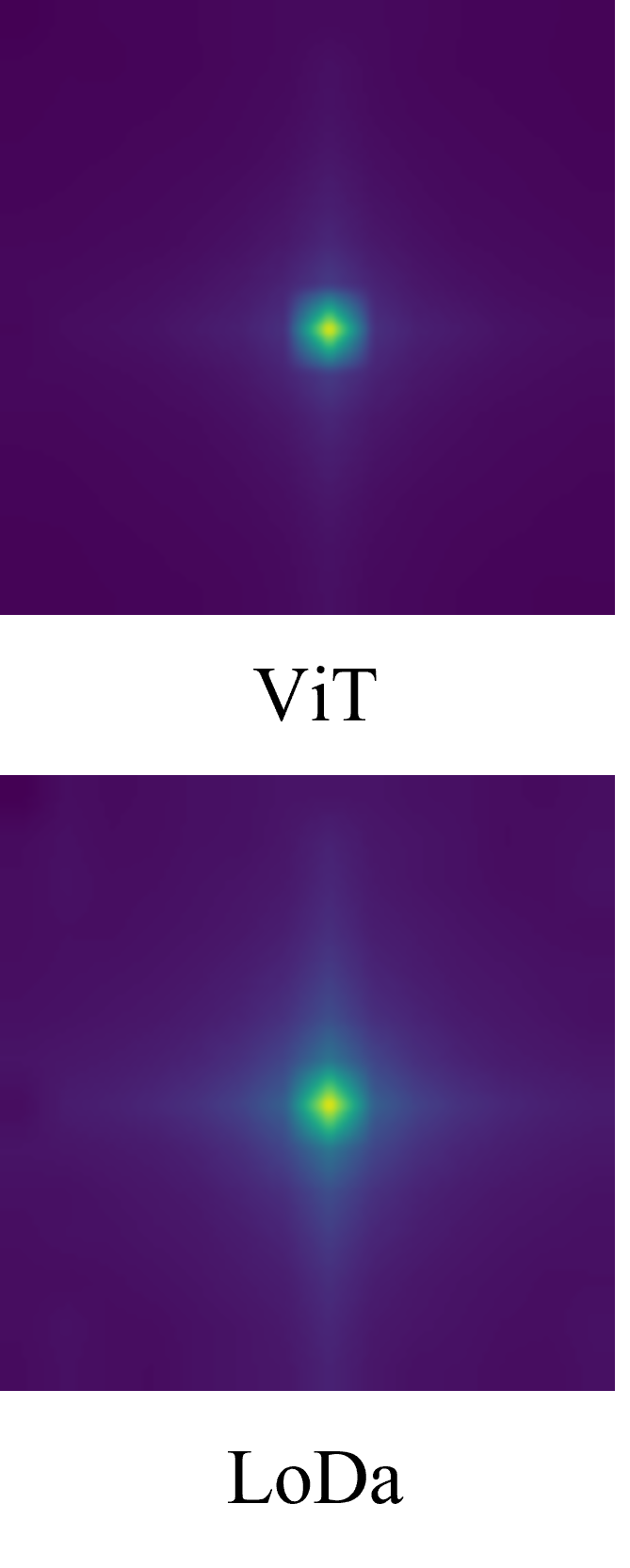}
        \caption{}
    \end{subfigure}
    \begin{subfigure}{0.69\linewidth}
        \centering
        \includegraphics[width=\linewidth]{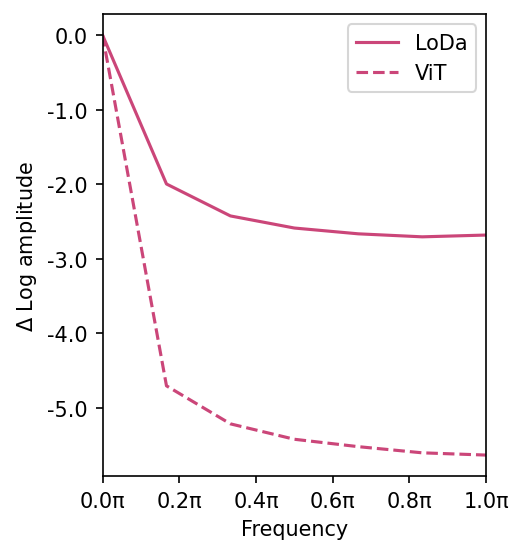}
        \caption{}
    \end{subfigure}
 \vspace{-2mm}
\caption{Fourier analysis of features of ViT and LoDa. (a) Fourier spectrum of ViT and LoDa. (b) Relative log amplitudes of Fourier Transformed feature maps. (a) and (b) show that LoDa captures more high-frequency signals.}
\label{fig:fourier}
 \vspace{-2mm}
\end{figure}

\subsection{Effectiveness of Large-scale Pretrained Models}
One of the key properties of our proposed methods is the efficient adaptation of large-scale pretrained models, which allows our model to achieve a competing performance to SOTA BIQA methods. To demonstrate the effectiveness of using large-scale pretrained models in our proposed models, we employ different pretrained weights, including ImageNet-1K pretrained weights~\cite{steinerHowTrainYour2022}, ImageNet-21K pretrained weights~\cite{steinerHowTrainYour2022}, MAE pretrained weights~\cite{heMaskedAutoencodersAre2022}, and Multi-Modal pretrained weights~\cite{chertiReproducibleScalingLaws2022}, and evaluate them on relatively large synthetical and authentical datasets, KADID-10k and KonIQ-10k. The experimental results are detailed in Table~\ref{tab:pre-train}. The transition from weights pretrained on ImageNet-1K to those pretrained on ImageNet-21K yields more benefits for our model, as the scale of pretraining data expansively increases. Besides, While MAE also employs ImageNet-1K pretraining, it distinguishes itself from supervised ImageNet-1K pretraining by embracing a more potent self-supervised pretraining approach, which also confers substantial advantages upon our model. However, our model faces challenges in effectively leveraging multi-modal pretrained weights. One plausible explanation is that multi-modal pretrained models may prioritize the abstract concepts inherent within images, a focus that diverges from the demands of IQA tasks. Since multi-modal pretrained weights contain more information than single-modal pretrained ones, how to apply these models to the IQA tasks will also be an important topic and we will commit to conducting further research on this. 

Moreover, the parameter capacities of large-scale pretrained models are another essential component of our method. To verify the effectiveness of large-scale pretrained model size, we evaluate LoDa with ViT-Tiny/Small/Base, and all ViTs are pretrained with ImageNet-21k. Quantitative results are shown in Table~\ref{tab:model-size}. From this, we can observe that with the growth of pretrained backbone sizes, our model can benefit from it and thus achieve better performance. In particular, solely employing ViT-S as the backbone, our method can achieve performance on par with SOTA shown in Table~\ref{tab:sota}, which further shows the effectiveness of our method.

\begin{table}[t]
    \centering
    \footnotesize
 \renewcommand\tabcolsep{7pt}
	\resizebox{\columnwidth}{!}{ 
    \begin{tabular}{c|cc|cc}
    \toprule
     & \multicolumn{2}{c|}{KADID-10k} & \multicolumn{2}{c}{KonIQ-10k}  \\
    Module & SRCC & PLCC & SRCC & PLCC \\
    \midrule
    ViT & 0.889 & 0.899 & 0.874 & 0.891 \\
    ViT + Extractor & 0.915 & 0.921 & 0.925 & 0.936 \\
    LoDa & \textbf{0.931} & \textbf{0.936} & \textbf{0.932} & \textbf{0.944} \\
    \bottomrule
    \end{tabular}}
 \vspace{-2mm}
    \caption{Ablation experiments on KADID-10K and KonIQ-10K datasets. Bold entries indicate the best results.}
    \label{tab:ablation}
    \vspace{-0.3cm}
\end{table}

\subsection{Comparisons with Different Fine-tuning Methods}
At present, numerous efficient model adaptation methods for large-scale pretrained vision models have emerged, Adapter~\cite{houlsbyParameterEfficientTransferLearning2019}, LoRA~\cite{huLoRALowRankAdaptation2022} and visual prompt tuning (VPT)~\cite{jiaVisualPromptTuning2022a} stands as the exemplars. To demonstrate the effectiveness of our proposed method, we employ linear probing ViT that only fine-tunes the head of ViT, full fine-tuning ViT, Adapter-ViT, LoRA-ViT, and VPT-ViT for IQA task, and compare it with our method on KADID-10K and KonIQ-10k datasets. The experimental results are detailed in Table~\ref{tab:ema}. It can be noticed that our model outperforms all of these fine-tuning methods on KADID-10K and KonIQ-10K, especially on KADID-10K, which shows the effectiveness and superiority of fusing CNN multi-scale distortion features into ViT.

\subsection{Ablation Study}
\subsubsection{Effect of CNN Features.}
Recent research~\cite{siInceptionTransformer2022} highlights the distinct characteristics exhibited by ViT and CNN. Specifically, it demonstrates that ViT is adept at learning low-frequency global signals, whereas CNN exhibits a propensity for extracting high-frequency information. Following previous work~\cite{siInceptionTransformer2022}, we visualize the Fourier analysis of features of ViT and our models (average over 128 images) in Figure~\ref{fig:fourier}. From the Fourier spectrum and relative log amplitudes of Fourier transformed feature maps, we can deduce that our model captures more high-frequency signals than the full-finetuned ViT baseline. And from Table~\ref{tab:ema}, we can also observe that our model outperforms the full-finetuned ViT baseline by a large margin. This enhanced capability can be attributed to the incorporation of fused multi-scale distortion features extracted by CNN.

\subsubsection{Ablation for Components.}
Our models is composed of three essential components, including the large-scale pretrained ViT, local distortion extractor, and local distortion injector. To examine the individual contribution of each component, we report the ablation experiments in Table~\ref{tab:ablation}. From this table, we observe that both the local distortion extractor and local distortion injector are highly effective in characterizing the image quality, and thus contributing to the overall performance of LoDa. In particular, even without local distortion injector, we simply add the multi-scale distortion tokens with ViT tokens, it can still outperform the full-finetuned ViT, demonstrating the effectiveness of adaptation of large-scale pretrained models and the extracted multi-scale distortion features. 

\section{Conclusion}
In this paper, we present a LOcal Distortion Aware efficient transformer adaptation (LoDa) for image quality assessment, which utilizes large-scale pretrained models. Given that IQA is highly reliant on both local and non-local dependencies, while ViT primarily captures the non-local aspects of images, overlooking the local details, henceforth, we propose the integration of CNN for extracting multi-scale distortion features and injecting them into ViT. However, for ViT extracts \(16 \times 16\) patches of images, directly adding these multi-scale distortion features to ViT tokens may encounter a challenge of misaligned scale. Thus we propose to utilize the cross-attention mechanism to let ViT tokens query related features from multi-scale distortion features and then combine them.  Experiments on seven standard datasets demonstrate the superiority of LoDa in terms of prediction accuracy, training efficiency, and generalization capability.

\section{Appendix}
\subsection{Introduction}
This supplementary material presents: (1) additional experimental analysis and quantitative results of the ablation study of LoDa; (2) more visualization of Fourier analysis of vision transformer (ViT)~\cite{dosovitskiyImageWorth16x162021} and LoDa. 

\subsection{More Ablation Study and Discussion}
\subsubsection{Evaluation Metrics}
The detailed definitions of the two performance metrics (\textit{i.e.}, SRCC, PLCC) we use in this paper are as follows:
\begin{equation}
    SRCC = 1 - \frac{6 \sum_{t=1}^T d_t}{T(T^2 - 1)}
\end{equation}
where \(T\) is the number of distorted images, and \(d_t\) is the rank difference between the ground-truth quality score and the predicted score of image \(t\). 
\begin{equation}
    P L C C=\frac{\sum_{t=1}^{T}\left(s_{t}-\bar{s}_{t}\right)\left(\hat{s}_{t}-\bar{\hat{s}}_{t}\right)}{\sqrt{\sum_{t=1}^{N}\left(s_{t}-\bar{s_{t}}\right)^{2}} \sqrt{\sum_{t=1}^{N}\left(\hat{s_{t}}-\bar{\hat{s}}_{t}\right)^{2}}}
\end{equation}
where \(\bar{s_{t}}\) and \(\bar{\hat{s}}_{t}\) denote the means of the ground truth and predicted score, respectively. 

\subsubsection{Latent Dimension}
Due to the potentially overwhelming number of parameters and computational overhead caused by the large dimension of ViT~\cite{dosovitskiyImageWorth16x162021} (768 for ViT-B), inspired by the concept of adapters in the field of NLP~\cite{houlsbyParameterEfficientTransferLearning2019}, we propose to down project the ViT tokens and multi-scale distortion tokens to a smaller latent dimension \(r\). We study the effect of the latent dimension \(r\) on KonIQ-10k~\cite{hosuKonIQ10kEcologicallyValid2020} and KADID-10k~\cite{linKADID10kLargescaleArtificially2019} datasets. Results are shown in Table~\ref{tab:latent-dim}. From the table, we can observe that on the KonIQ-10k dataset, our model is slightly affected by the effect of latent dimension \(r\), and on the KADID-10k dataset, our model performs the best when \(r\) is 64. Therefore, we empirically set \(r\) to 64 by default.

\begin{table}[t]
    \centering
    \footnotesize
 \renewcommand\tabcolsep{9pt}
	\resizebox{\columnwidth}{!}{ 
    \begin{tabular}{c|cc|cc}
    \toprule
     & \multicolumn{2}{c|}{KADID-10k} & \multicolumn{2}{c}{KonIQ-10k}  \\
    \(r\) & SRCC & PLCC & SRCC & PLCC \\
    \midrule
    48 & 0.929 & 0.934 & \textbf{0.934} & \textbf{0.945} \\
    64 & \textbf{0.931} & \textbf{0.936} & 0.932 & 0.944 \\
    80 & 0.923 & 0.928 & 0.933 & \textbf{0.945} \\
    \bottomrule
    \end{tabular}}
 \vspace{-2mm}
    \caption{Impact of the latent dimension \(r\). The best performances are highlighted with \textbf{boldface}.}
    \label{tab:latent-dim}
    \vspace{-0.3cm}
\end{table}

\subsubsection{Number of Heads in Cross-attention}
We run an ablation study on different numbers of heads in cross-attention when the latent dimension is set to 64. As shown in Table~\ref{tab:head-num}, when the latent dimension is fixed, the number of heads in cross-attention has little effect on our model. Thus, we set the number of heads to four so that our model performs slightly better on the KADID-10k dataset. 

\begin{table}[t]
    \centering
    \footnotesize
 \renewcommand\tabcolsep{9pt}
	\resizebox{\columnwidth}{!}{ 
    \begin{tabular}{c|cc|cc}
    \toprule
     & \multicolumn{2}{c|}{KADID-10k} & \multicolumn{2}{c}{KonIQ-10k}  \\
    \(h\) & SRCC & PLCC & SRCC & PLCC \\
    \midrule
    2 & 0.929 & 0.934 & 0.932 & 0.944 \\
    4 & \textbf{0.931} & \textbf{0.936} & \textbf{0.932} & \textbf{0.944} \\
    8 & 0.929 & 0.935 & 0.933 & 0.944 \\
    \bottomrule
    \end{tabular}}
 \vspace{-2mm}
    \caption{Impact of the number of heads \(h\) in cross-attention.}
    \label{tab:head-num}
    \vspace{-0.3cm}
\end{table}

\begin{figure}[t]
\centering
    \begin{subfigure}{0.29\linewidth}
        \centering
        \includegraphics[width=\linewidth]{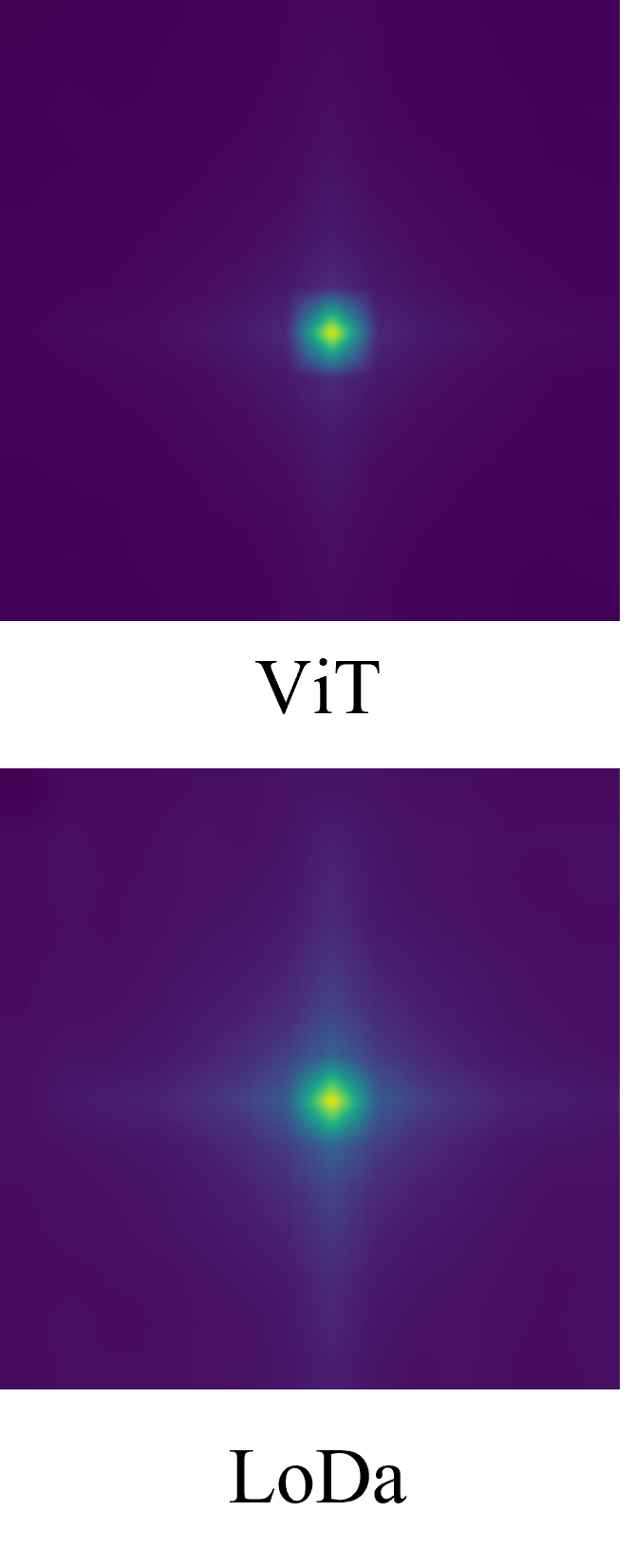}
        \caption{}
    \end{subfigure}
    \begin{subfigure}{0.69\linewidth}
        \centering
        \includegraphics[width=\linewidth]{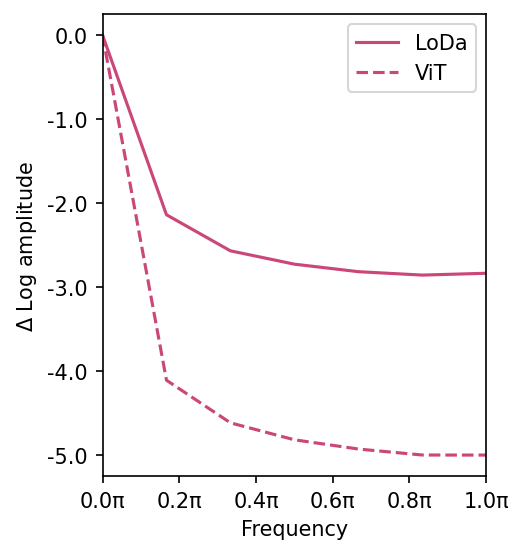}
        \caption{}
    \end{subfigure}
 \vspace{-2mm}
\caption{Fourier analysis of features of ViT and LoDa on KonIQ-10k. (a) Fourier spectrum of ViT and LoDa. (b) Relative log amplitudes of Fourier Transformed feature maps. (a) and (b) show that LoDa captures more high-frequency signals.}
\label{fig:fourier_koniq10k}
 \vspace{-2mm}
\end{figure}

\subsubsection{Number of Interactions}
In the paper, we empirically fuse the ViT tokens with multi-scale features in all of the ViT encoder layers. However, it's essential to acknowledge that this choice is a result of our empirical decision-making process, in fact, we can fuse them only in part of ViT encoder layers. Thus, we run an ablation study with different numbers of interactions \(N\)on KonIQ-10k and KADID-10k datasets. In this ablation study, we distribute the ViT encoder layers into \(N\) blocks, with each block containing \(L / N\) encoder layers, where \(L\) denotes the total number of encoder layers. Then, we only fuse the ViT tokens with multi-scale distortion features in each block instead of each layer. Results are shown in~\ref{tab:interaction-num}. It can be observed that our model's performance improves with an increased number of interactions. Notably, it is worth observing that even with just half of the interactions, our model yields excellent performance outcomes.

\begin{table}[t]
    \centering
    \footnotesize
 \renewcommand\tabcolsep{9pt}
	\resizebox{\columnwidth}{!}{ 
    \begin{tabular}{c|cc|cc}
    \toprule
     & \multicolumn{2}{c|}{KADID-10k} & \multicolumn{2}{c}{KonIQ-10k}  \\
    \(N\) & SRCC & PLCC & SRCC & PLCC \\
    \midrule
    3 & 0.923 & 0.929 & 0.929 & 0.941 \\
    6 & 0.927 & 0.933 & \textbf{0.932} & 0.943 \\
    12 & \textbf{0.931} & \textbf{0.936} & \textbf{0.932} & \textbf{0.944} \\
    \bottomrule
    \end{tabular}}
 \vspace{-2mm}
    \caption{Impact of the number of heads \(h\) in cross-attention.}
    \label{tab:interaction-num}
    \vspace{-0.3cm}
\end{table}

\subsection{Visualization of Fourier Analysis of Vision Transformer and LoDa}
In the paper, we show the Fourier analysis of features of ViT and LoDa on the KADID-10k dataset, here we additionally show the Fourier analysis of full-fintuned ViT and LoDa on the KonIQ-10k dataset (average over 128 images) in Figure~\ref{fig:fourier_koniq10k}. We can observe the same results on the KonIQ-10k dataset, it further demonstrates that LoDa captures more high-frequency signals and show the effect of our proposed method. 

\bibliographystyle{IEEEtran}
\bibliography{IEEEabrv,conference}

\end{document}